\documentclass[final,1p,times]{elsarticle}
\usepackage{color}
\pdfoutput=1
\usepackage{epsfig}
\usepackage{amsmath}
\usepackage{amssymb}
\usepackage{epstopdf}
\usepackage[center]{subfigure}
\usepackage{multirow}
\usepackage{graphicx}
\usepackage{algorithm}
\usepackage{algorithmic}
\usepackage{rotating}

\usepackage{multirow}
\usepackage{url}
\usepackage{caption}
\usepackage[pagewise]{lineno}\linenumbers

\newcommand{\RNum}[1]{\uppercase\expandafter{\romannumeral #1\relax}}

\usepackage{amssymb}

\journal{}

\begin{document}

\begin{frontmatter}



\title{WAD-CMSN:  Wasserstein Distance based Cross-Modal Semantic Network for Zero-Shot Sketch-Based Image Retrieval\tnoteref{fund}}
	
\tnotetext[fund]{The work described in this paper is supported partially by the National Natural Science Foundation of China (11871167),  Project of Guangdong Province Innovative Team (2017WCXTD004, 2020WCXTD011),   Foundation of Guangdong Educational Committee (2019KZDZX1023), Special Support Plan for High-Level Talents of Guangdong Province (2019TQ05X571).}

\author[label1]{Guanglong Xu}
\ead{guanglongxu2018@gmail.com}

\author[label2]{Zhensheng Hu}
\ead{huzhsh6@mail2.sysu.edu.cn}

\author[label3]{Jia Cai\corref{cor1}}
\ead{jiacai1999@gdufe.edu.cn}
\cortext[cor1]{Corresponding author}

\address{\fnref{label1}}
\fntext[label1]{School of Economics and Finance, South China University of Technology,  Guangzhou, 510641, P. R. China.}

\address{\fnref{label2}}
\fntext[label2]{School of Computer Science and Engineering, Sun Yat-Sen University,  Guangzhou, 510006, P. R. China.}

\address{\fnref{label3}}
\fntext[label3]{School of Artificial Intelligence and Digital Economy  Industry,  Guangdong University  of Finance $\&$ Economics,  Guangzhou, 510320, P. R. China.}

\begin{abstract}
Zero-shot sketch-based image retrieval (ZSSBIR), as a popular studied  branch of computer vision, attracts wide attention recently. Unlike sketch-based image retrieval (SBIR), the main aim of ZSSBIR is to retrieve natural images given free hand-drawn sketches that may not appear during training. Previous approaches used semantic aligned sketch-image pairs or utilized memory expensive fusion layer for projecting the visual information to a low dimensional subspace, which ignores the significant heterogeneous cross-domain discrepancy between highly abstract sketch and relevant image. This may yield poor performance in the training phase. To tackle this issue and overcome this drawback,  we propose a Wasserstein distance based cross-modal semantic network (WAD-CMSN) for ZSSBIR. Specifically, it first projects the visual information of each branch (sketch, image) to a common  low dimensional  semantic subspace via Wasserstein distance in an adversarial training manner. Furthermore, identity matching loss is employed to select useful  features, which can not only capture complete semantic knowledge, but also alleviate the over-fitting phenomenon caused by the WAD-CMSN model. Experimental results  on the challenging Sketchy (Extended) and TU-Berlin (Extended) datasets indicate the effectiveness of the proposed WAD-CMSN model over several competitors.
\end{abstract}
\end{frontmatter}
\section{Introduction}\label{sec:intr}
Image retrieval plays important roles in modern  science and technology, which has popular applications in computer vision,  medical image analysis, and e-commerce. Traditional image retrieval utilizes textual descriptions to retrieve images given queries, which may be difficult to achieve in real-world situations. Due to the rapid development and prosperity of touch screen devices,  free hand-drawn sketches based image retrieval attracts serious widespread attention, which formed a new area, namely, sketch-based image retrieval  (SBIR).  A sketch is a quickly-done drawing without many details, which is highly abstract and intuitive to human beings. In practice, we may confront with the difficulty of training all the classes, which is unrealistic  and may lead to poor retrieval performance when testing on unseen classes.  Hence, ZSSBIR, a generalization of SBIR in the zero-shot learning (ZSL) setting, have emerged in real-world applications. ZSSBIR is an extremely challenging task, which needs to simultaneously  tackle the problems of  cross-modal matching, heterogeneous domain discrepancy and limited knowledge of the unseen classes.  Traditional SBIR methods may face the over-fitting problem in the source domain, neglect the unseen classes and cannot handle the above issues. Meanwhile, conventional ZSL methods can only deal with the single modality problems. Thus, ZSSBIR tries to combine the advantages of ZSL and SBIR. Prior works on ZSSBIR attempt to learn a mapping from an input sketch to a relevant image using labelled aligned pairs, which is either unavailable or very expensive in practice. Furthermore, the introduction of fusion layer that jointly represents two or more modalities is costly in terms of memory.  Semantically paired   cycle consistent generative  model \citep{dutta2019semantically} was delicate designed to alleviate these drawbacks, while progressive independent feature decomposition cross-modal network (PDFD) \citep{xu2020progressive} was also investigated. However, as demonstrated in \cite{arjovsky2017towards}, \citep{dutta2019semantically} and \citep{xu2020progressive} try to minimize the  Jensen-Shannon (JS) divergence  of the two heterogeneous distributions, which may be inappropriate. JS divergence can only portray the situation when the two distributions have overlapping areas. If  the two distributions have almost no intersection, the gradient  of the JS divergence is zero, which means the parameters of the model cannot be updated. However,  the Wasserstein distance can alleviate these shortcomings due to the fact that sketch and image are from heterogeneous domains with different distributions. 
	
In this article, we propose a Wasserstein distance based cross-modal semantic network (WAD-CMSN) for the ZSSBIR task. The main idea can be addressed as follows. Firstly, each branch maps the features of sketch (image)  to a common  low dimensional semantic subspace based on Wasserstein distance  via adversarial training. Secondly, the cycle consistency constraint is introduced on each branch. On one hand, it enforces to project the sketch and image modality to a common subspace. On the other hand, it can be guaranteed that we can reach  to the original sketch or image modality when translating the features back to the original modality.  In addition, classification constraint on the generators' output will produce discriminative generated features.  Thirdly, imposing by the label embedding method \citep{akata2015evaluation}, we utilize text-based and hierarchical models to conduct semantic knowledge supervision. The main contributions  are described as the following.
\begin{itemize}
	\item We present a novel  Wasserstein distance  based adversarial loss and identity matching loss to efficiently bridge the heterogeneous domain gap between sketch and image modality.
	
	\item The proposed approach can well capture the complete semantic features of sketch-image pairs, mitigate the over-fitting phenomenon.
	
	\item Extensive experiments are investigated on two large-scale datasets, namely, Sketchy (Extended) and TU-Berlin (Extended)  to indicate the efficiency and effectiveness of the WAD-CMSN model.
\end{itemize}
The remainder of the paper is organized as follows. In Section \ref{section2}, we give a brief review of SBIR, ZSL and ZSSBIR. Section \ref{section3} devotes to the design of the novel algorithm. Section \ref{section4}  discusses the experimental results. Section \ref{section5} concludes the paper.
\section{Related Work}\label{section2}
Firstly, let us give a brief review of  SBIR, ZSL and ZSSBIR.

{\bf Sketch-Based Image Retrieval}.
The main aim of SBIR is to reduce the cross-domain gap between sketches and relevant images, which consists of two categories: hand-crafted based methods and deep learning-based methods \citep{liu2017deep}. Hand-crafted approaches try to extract the edge map of images, then match those images with highly abstract sketch. Famous methods such as gradient field HOG descriptor \citep{hu2013performance}, Fuzzy rule \citep{khan2019fuzzy}, histogram of oriented edges,  \citep{saavedra2014sketch}, and learned key shapes \citep{saavedra2015sketch}  fall into this category. For deep learning-based methods,  \cite{yu2016sketch} used  CNN  and developed a deep model based on triplet ranking loss to tackle SBIR for the first time. Siamese networks \citep{chopra2005learning} are found suitable for cross-modal retrieval. Distinct type of loss functions such as  contrastive loss \citep{chopra2005learning}, triplet ranking loss \citep{sangkloy2016sketchy, yu2016sketch}, and higher-order learnable energy function (HOLEF) based loss \citep{song2017deep} are employed  and investigated in these frameworks. Xu et al. \cite{xu2018cross} proposed a novel cross-modality representation learning paradigm for SBIR.  The above-mentioned references focus on coarse-grained SBIR. However, fine-grained SBIR gained widespread popularity recently. A novel deep neural network, namely, sketch-a-net \citep{yu2017sketch} was designed for instance-level SBIR issue.

{\bf Zero-Shot Learning}.
Zero-shot learning is a recently widely discussed branch of computer vision, which refers to retrieve objects that are not available during the training phase. Generally, it consists of two  categories, viz, embedding-based and generative-based approaches. Embedding-based approaches learn nonlinear multi-modal embedding \citep{akata2015label,changpinyo2017predicting,socher2013zero,xian2016latent,zhang2017learning}. For generative-based approaches, \cite{bucher2017generating} introduced a conditional generative moment matching network to synthesize the features of unseen categories. Chen et al. \cite{chen2018zero} investigated semantic-preserving adversarial embedding network for ZSL. ZSL requires side information, which can effectively transfer learned knowledge  from seen categories  to unavailable categories. Attribute is one famous type of side information, which requires expensive human annotation, however. Thus, a flurry of works \citep{akata2015evaluation, xian2016latent,reed2016learning} utilized text-based model \citep{mikolov2013efficient} or hierarchical model \citep{miller1995wordnet} as auxiliary information for label embedding.

{\bf Zero-Shot Sketch-Based Image Retrieval}.
Shen et al. \cite{shen2018zero} investigated a generative hashing model to solve the ZSSBIR task for the first time.  Yelamarthi et al. \cite{yelamarthi2018zero} proposed conditional variational and conditional adversarial auto-encoder based models for the ZSSBIR task. Recently,  \cite{dutta2019semantically} utilized category-level side information and disclosed a paired cycle consistent generative model (SEM-PCYC). Moreover, \cite{dutta2019style} proposed a novel style-guided fake-image generation method, while  \cite{deng2020progressive} addressed a progressive  semantic network by first aligning the features of sketch and image, and subsequently projecting the aligned features to a common low dimensional subspace.
\section{The proposed approach}\label{section3}
\begin{figure}[htb]
	\centering
	\includegraphics[scale=0.6]{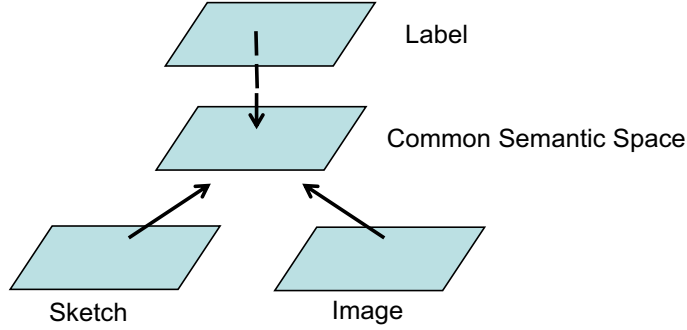}
	\caption{The pipeline of SBIR, which simultaneously projects sketch/image features and label information into common  low dimensional semantic space for efficient retrieval.}
	\label{pipeline}
\end{figure}

\begin{figure*}[htb]
	\centering
	\includegraphics[scale=0.8]{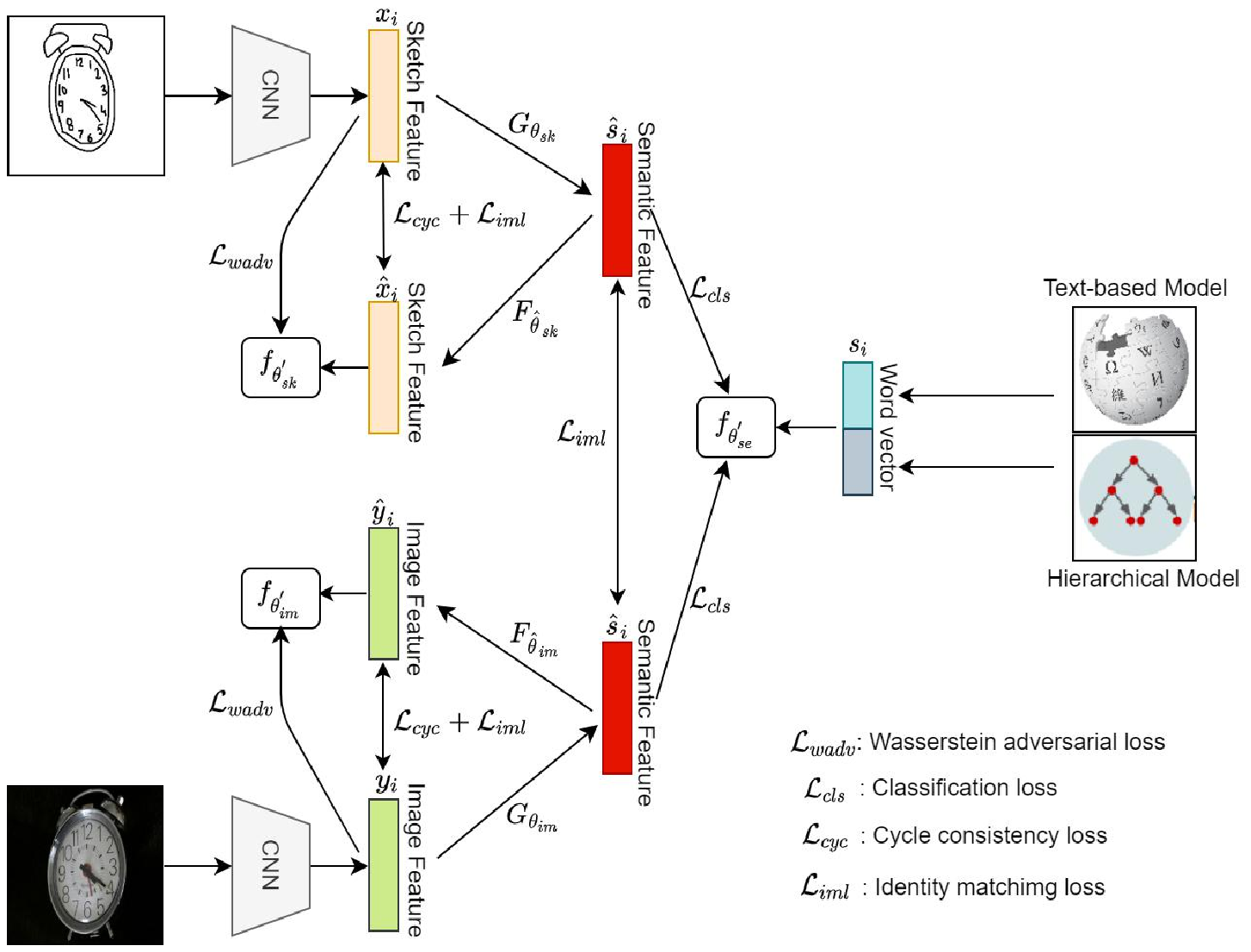}
	\caption{Structure of WAD-CMSN.}
	\label{Structure}
\end{figure*}

In this paper, we address a two-branch Wasserstein distance based cross-modal semantic network, which embodies the discrepancy of different domains (sketch, image). The proposed model projects sketch and image features to a common semantic subspace in an adversarial manner under the semantic knowledge supervision. Before depicting the proposed WAD-CMSN model, let us first show the idea of SBIR task (Fig.\ref{pipeline}). The structure of SBIR contains two branches for sketch and image, respectively. Let ${\cal D}_{tr}$ $=$ $\{X^{tr}$, $Y^{tr}$, $S^{tr}$, $C^{tr}\}$ be a training set contains $N_s$ samples of  sketches $X^{tr}=\{x_i^{tr}\}^{N_s}_{i=1}$, images $Y^{tr}=\{y_i^{tr}\}^{N_s}_{i=1}$, semantic embedding $S^{tr}=\{s_i^{tr}\}^{N_s}_{i=1}$  and category labels $C^{tr}=\{c_i^{tr}\}^{N_s}_{i=1}$. Let us denote  ${\cal D}_{ts}=\{X^{ts}, Y^{ts},  C^{ts}\}$ as the test set, which contains sketches  $X^{ts}$ $=$ $\{x_i^{ts}\}$$^{N_u}_{i=1}$, images $Y^{ts}$$=$$\{y_i^{ts}\}^{N_u}_{i=1}$,  and category labels $C^{ts}=\{c_i^{ts}\}^{N_u}_{i=1}$ with $N_u$ samples. Since we consider SBIR in the zero-shot setting, obviously, $C^{tr}\bigcap C^{ts}= \emptyset$. Thus, $X^{tr}$ and $Y^{tr}$  are used for training. At the test phase, given an unseen sketch $x^{ts}_i$ in $X^{ts}$, ZSSBIR aims at retrieving  images from the  gallery $Y^{ts}$. The objective of WAD-CMSN model is to learn the discriminative function $f_{\theta'}$, and generators $G_{\theta_{sk}}(\cdot): \mathbb{R}^d\to \mathbb{R}^M$, $G_{\theta_{im}}(\cdot): \mathbb{R}^d\to \mathbb{R}^M$ for sketch and image, respectively. Then project them into a common semantic space, where the learned knowledge could be transferred to the unavailable classes. The structure of the model could be depicted in Fig. \ref{Structure}.

\subsection{Semantic feature generation}
In practice, sketch and image are not aligned, which results in knowledge loss. To alleviate this kind of problem, semantic embedding is introduced, which can provide knowledge supervision at category levels, align the projected features of sketch and image to the word vector in an  adversarial manner.

{\bf Semantic knowledge embedding}:  Text-based and hierarchical models are widely utilized to attain initial side information. Word2vec and Glove are two famous important text-based embeddings. Word2vec,  a two layered neural network, can map  words into a high dimensional vector space, where each word is represented by a vector.  Glove learns word vectors based on word co-occurrence matrix, which can effectively encode crucial semantic information, while WordNet is utilized to attain  semantic similarity for hierarchical model due to the fact that most of classes of Sketchy (Extended) and TU-Berlin (Extended) are contained in WordNet. Thus, the similarity between words can be measured by path similarity (computing the number of hops) and Jiang-Conrath \citep{jiang1997semantic}. Moreover, the text-based and hierarchical models are combined via a feature selection auto-encoder \cite{wang2017}.

As mentioned above, sketch and image features have significant domain gap. To efficiently bridge the heterogeneous domain gap, an appropriate distance metric is necessary. In the literature, Wasserstein distance is suitable to measure the domain discrepancy. Recall the Wasserstein distance (also called earth mover's distance, EM  distance) between the source and target domains is computed as the following:
\begin{equation}\label{wdis}
W(p_r,p_{\bf z})= \inf_{\gamma\in \Pi (p_r, p_{\bf z})} {\mathbb E}_{(x,y)\in\gamma} \big[\|x-y\|\big],
\end{equation}
where  $p_r$ and $p_{\bf z}$  denote two distributions, and $\Pi (p_r, p_{\bf z})$ is the set of all joint distributions  $\gamma (x,y)$. In \cite{arjovsky2017wasserstein}, it has been proved that Wasserstein distance is more suitable to measure the discrepancy of two distinct distributions than KL divergence and JS divergence. 

{\bf Wasserstein distance based adversarial loss}: Ref. \cite{arjovsky2017wasserstein} stated that the infimum in Eq.  (\ref{wdis}) is intractable. Thus, to tackle the feature gap between sketches and images,  we introduce the following Wasserstein distance based adversarial loss inspired from Eq. (2) in \cite{arjovsky2017wasserstein}:
$$\min_G\max_{\|f\|_L\leq 1} {\cal L}(f,G)= \mathbb {E}_{{\bf x}\sim p_r({\bf x})} [f({\bf x})]-\mathbb {E}_{{\bf z}\sim p_{\bf z}({\bf z})}[f(G({\bf z}))],$$
where $\|f\|_L\leq 1$ means the $1-$Lipschitz function.
As indicated in Fig. \ref{Structure}, each branch (sketch or image) contains a generator $G_\theta$ and a common discriminative function $f_{\theta'}$. The features of  sketch-image pairs are extracted utilizing  pre-trained VGG16 \citep{simonyan2014very}  on ImageNet  \citep{deng2009imagenet} database.  We consider four generators $G_{\theta_{sk}}: X\to S$, $G_{\theta_{im}}: Y\to S$, $F_{\hat \theta_{sk}}:S\to X$ and $F_{\hat\theta_{im}}: S\to Y$. $G_{\theta_{sk}}$ and $G_{\theta_{im}}$ try to minimize the objective stated in Eq. (\ref{wad-loss1}) against $f_{\theta'_{se}}$ that attempts to maximize it,
\begin{equation}\label{wad-loss1}
\min_{G_{\theta_{sk}}, G_{\theta_{im}}} \max_{f_{\theta'_{se}}} {\cal L}_{wadv} (G_{\theta_{sk}}, G_{\theta_{im}}, f_{\theta_{se}}, {\bf x}, {\bf y}, {\bf s}^{tr}) = 2\mathbb{E}[f_{\theta'_{se}}( {\bf s}^{tr})]-\mathbb {E}[f_{\theta'_{se}}(G_{\theta_{sk}}({\bf x}))]-\mathbb {E}[f_{\theta'_{se}}(G_{\theta_{im}}({\bf y}))],
\end{equation}

where $ {\bf x}$ (${\bf y}$ resp.) is the sketch (image resp.) feature, ${\bf s}^{tr}$ denotes the   word vector, $G_{\theta_{sk}}$ ($G_{\theta_{im}}$ resp.) stands for sketch (image resp.) semantic generation function, while $f_{\theta'_{se}}$ means discriminative function.  Similarly, for the generator $F_{\hat \theta_{sk}}$ (resp. $F_{\hat \theta_{im}}$) and its discriminative function  $f_{\theta'_{sk}}$ (resp. $f_{\theta'_{im}}$), the objective is
\begin{eqnarray}
\label{wad-loss2}
&&\min_{F_{\hat\theta_{sk}}} \max_{f_{\theta'_{sk}}} {\cal L}_{wadv} (F_{\hat\theta_{sk}}, f_{\theta'_{sk}},  {\bf x}, {\bf s}^{tr})
= \mathbb{E}[ f_{\theta'_{sk}}({\bf x})]-\mathbb {E}[f_{\theta'_{sk}}(F_{\hat\theta_{sk}}({\bf s}^{tr}))],
\end{eqnarray}
and
\begin{eqnarray}
\label{wad-loss3}
&&\min_{F_{\hat\theta_{im}}}  \max_{f_{\theta'_{im}}}  {\cal L}_{wadv} (F_{\hat\theta_{im}}, f_{\theta'_{im}}, {\bf y}, {\bf s}^{tr})
 = \mathbb{E}[ f_{\theta'_{im}}({\bf y})] -\mathbb {E}[f_{\theta'_{im}}(F_{\hat\theta_{im}}({\bf s}^{tr}))],
\end{eqnarray}
where $F_{\hat\theta_{sk}}$ ($F_{\hat\theta_{im}}$ resp.) minimize the objective and $f_{\theta'_{sk}}$ ($f_{\theta'_{im}}$ resp.) tries to maximize it.
Therefore, the main aim of  Wasserstein distance based adversarial loss is to learn semantic features that are expected to be similar to the word vectors.

{\bf Cycle consistency loss}:  Previous adversarial approach could bridge the significant cross-domain gap efficiently. However, there is no guarantee whether an input and its output are well matched. Similar to \citep{dutta2019semantically}, we consider cycle consistency  constraint.  When we project the features of a sketch (resp. image) to the  low dimensional  semantic subspace, and then interpret it back to the sketch (resp. image) feature space, we should arrive back at the same original sketch (resp. image) features. Therefore, cycle consistency loss helps learning mappings across domains when we can not obtain aligned samples.  The cycle consistency loss can be defined as
\begin{eqnarray}
\label{cyc-loss1}
&&\min_{G_{\theta_{sk}}, F_{\hat\theta_{sk}}}{\cal L}_{cyc}(G_{\theta_{sk}}, F_{\hat\theta_{sk}})=\mathbb {E} [\|F_{\hat\theta_{sk}}(G_{\theta_{sk}}({\bf x}))-{\bf x}\|_1]
+\mathbb {E} [\|G_{\theta_{sk}}(F_{\hat\theta_{sk}}({\bf s}^{tr}))-{\bf s}^{tr}\|_1],
\end{eqnarray}
\begin{eqnarray}
\label{cyc-loss2}
&&\min_{G_{\theta_{im}}, F_{\hat\theta_{im}}}{\cal L}_{cyc}(G_{\theta_{im}}, F_{\hat\theta_{im}})=\mathbb {E} [\|F_{\hat\theta_{im}}(G_{\theta_{im}}({\bf y}))-{\bf y}\|_1]
+\mathbb {E} [\|G_{\theta_{im}}(F_{\hat\theta_{im}}({\bf s}^{tr}))-{\bf s}^{tr}\|_1],
\end{eqnarray}

{\bf Classification loss}:  Obviously, adversarial training presented in Eqs.(\ref{wad-loss1})-(\ref{wad-loss3}) and cycle consistency constraints outlined in Eqs. (\ref{cyc-loss1}) and (\ref{cyc-loss2}) do not ensure whether we can obtain discriminative generated features. To overcome this drawback, we add discriminative classifier  on the input data, and consider
\begin{align}
\label{cls-loss}
\min_{G_{\theta_{sk}}}{\cal L}_{cls} (G_{\theta_{sk}})= - \mathbb {E} [\log P(c|G_{\theta_{sk}} ({\bf x}))],\notag\\
\min_{G_{\theta_{im}}}{\cal L}_{cls} (G_{\theta_{im}})= - \mathbb {E} [\log P(c|G_{\theta_{im}} ({\bf y}))],
\end{align}
where $c$ is the category label of ${\bf x}$ and ${\bf y}$.

{\bf Identity matching loss}: Since we extract features from  different categories, it is difficult to measure the discrepancy between the transformed  features  and those before  transformed. For a given category, the semantic features of a sketch and its corresponding  image are the same, i.e., $G_{\theta_{sk}} ({\bf x})$$ \approx$ $G_{\theta_{im}} ({\bf y})$. On the contrary, under the projection of generator $F(\cdot)$, the features of a sketch or an image should be similar to the semantic features.
Hence, we need to minimize
\begin{equation}
\label{iml-loss}
{\cal L}_{iml} (G_{\theta_{sk}}, G_{\theta_{im}}, F_{\hat\theta_{sk}}, F_{\hat\theta_{im}})= \|G_{\theta_{sk}} ({\bf x})- G_{\theta_{im}} ({\bf y})\|^2_2+\|F_{\hat\theta_{sk}}({\bf s}^{tr})-{\bf x}\|^2_2+\|F_{\hat\theta_{im}}({\bf s}^{tr})-{\bf y}\|^2_2,
\end{equation}
with respect to (w.r.t) $G_{\theta_{sk}}$, $G_{\theta_{im}}$, $F_{\hat \theta_{sk}}$, and $F_{\hat\theta_{im}}$. The identity matching loss introduced here is different from the  previously discussed identity loss and cycle consistency loss. On one hand, we use $\ell_2$ norm  instead of $\ell_1$ norm, since $\ell_1$ norm  may lead to sparse results and yield  poor performance in the final model. On the other hand, whether the projected ${\bf x}$ and ${\bf y}$ in the semantic space, the  input ${\bf x}$ ($\bf y$) and the mapping of ${\bf s}^{tr}$ are well-matched  cannot be guaranteed by the cycle consistency loss and the identity loss. Thus, based upon the mentioned above steps, we can extract semantic features from sketch and real image through adversarial training to achieve cross-domain retrieval.
\subsection{Objective and Optimizer}
Therefore, the full model can be addressed as follows:
\begin{equation}
\label{total-loss}
{\cal L}_{total}=  {\cal L}_{wadv}+ {\cal L}_{cyc}+ {\cal L}_{cls}+ {\cal L}_{iml},
\end{equation}
 We train  Eq. (\ref{total-loss}) using  RMSprop in Pytorch, and depict the optimization mechanism in  Algorithm \ref{WAD-CMSN}. The expression of ${\cal L}_{wadv}$ indicates that the generators $G_{\theta_{sk}}$, $G_{\theta_{im}}$, $F_{\hat\theta_{sk}}$ and $F_{\hat\theta_{im}}$ minimize ${\cal L}_{wadv}$ against discriminative functions $f_{\theta'_{se}}$, $f_{\theta'_{sk}}$, and $f_{\theta'_{im}}$ that try to maximize ${\cal L}_{wadv}$. We can decompose ${\cal L}_{wadv}= {\cal L}_{gen}+ {\cal L}_{dis}$, where ${\cal L}_{gen}$ denotes the loss of the generators, and ${\cal L}_{dis}$ stands for the loss of the discriminative functions. In this paper, we first update the discriminators' parameters with respect to ${\cal L}_{dis}$, and then optimize the network parameters of the WAD-CMSN model with respect to ${\cal L}_{ps}$:
\begin{equation}
\label{pssum}
{\cal L}_{ps}=  {\cal L}_{gen}+  {\cal L}_{cyc}+  {\cal L}_{cls}+ {\cal L}_{iml}.
\end{equation}

\begin{algorithm}[!h]
	\renewcommand{\algorithmicrequire}{\textbf{Input:}}
	\renewcommand\algorithmicensure {\textbf{Output:} }
	\caption{WAsserstein Distance based Cross-Modal Semantic Network (WAD-CMSN).}
	\label{WAD-CMSN}
	\begin{algorithmic}[1]
		\REQUIRE ~~\\
		Data ${\cal D}_{tr}=\{X^{tr}, Y^{tr}, S^{tr}, C^{tr}\}$, max training iteration $M$, batch size $BN$.\\
		\ENSURE ~~\\
		$\theta_{sk}$, $\theta_{im}$, ${\hat \theta}_{sk}$, ${\hat \theta}_{im}$, $\theta'_{sk}$, $\theta'_{se}$,  $\theta'_{im}$\\
		\STATE Initial parameters $\theta_{sk}$, $\theta_{im}$, ${\hat \theta}_{sk}$, ${\hat \theta}_{im}$, $\theta'_{sk}$, $\theta'_{se}$,  $\theta'_{im}$,  ${\bf s}^{tr}$;\\
		\FOR {$t=1$ to $M$ }
		\STATE Generate sketch and image features ${\bf x}$, ${\bf y}$;\\
		\STATE Compute Wasserstein distance based adversarial loss via Eqs. (\ref{wad-loss1}), (\ref{wad-loss2}), and (\ref{wad-loss3});\\
		\STATE Compute cycle consistency loss via Eqs. (\ref{cyc-loss1}) and  (\ref{cyc-loss2});\\
		\STATE Compute classification loss via Eq. (\ref{cls-loss});\\
		\STATE Compute identity matching loss via Eq. (\ref{iml-loss});\\
		\STATE Update $\theta_{sk}$, $\hat\theta_{sk}$, $\theta_{im}$, $\hat\theta_{im}$ via stochastic gradient descent  w.r.t. ${\cal L}_{dis}$;\\
		\STATE Compute ${\cal L}_{ps}$ via Eq. (\ref{pssum});\\
		\STATE Update $\theta'_{sk}$, $\theta'_{se}$, $\theta'_{im}$ via stochastic gradient descent w.r.t. ${\cal L}_{ps}$;\\
		\ENDFOR
		\RETURN $\theta_{sk}$, $\theta_{im}$, ${\hat \theta}_{sk}$, ${\hat \theta}_{im}$, $\theta_{f_{sk}}$, $\theta_{f_{se}}$, $\theta_{f_{im}}$.\\
	\end{algorithmic}
\end{algorithm}
\section{Experiment}\label{section4}

\subsection{Setting}
There are two popular widely used benchmarks to evaluate the performance of SBIR, i.e., Sketchy (Extended) and TU-Berlin (Extended). We will conduct experiments on them.

{\bf Sketchy} is a large collection of sketch-photo pairs with $125$ categories. It contains $75,479$ sketches and $12,500$ images.  Liu et al. \cite{liu2017deep} extended the original Sketchy dataset by adding extra $60,502$ images. Hence, Sketchy (Extended) has totally $73,002$ images. We select $100$ seen classes for training while the remaining $25$ categories are used for test.

{\bf TU-Berlin} contains $20,000$ sketches and $204,489$ images from  $250$ categories. Unlike Sketchy (Extended) dataset, TU-Berlin dataset only has category-level matches. Following the data partition method outlined in \cite{dutta2019semantically}, we select the same $30$ categories  of sketches and images as a query set, while the remaining $220$ categories are used for training.

{\bf Baselines}:
We compare the proposed method with three ZSSBIR  methods:

\begin{itemize}
	\item \textbf{CVAE  \citep{yelamarthi2018zero}} introduced two auto-encoder models for  the ZSSBIR task, i.e., conditional variational based and contitional adversarial based models.
	
	\item \textbf{SEM-PCYC \citep{dutta2019semantically}}  presented a semantically aligned cycle-consistency generative model under the semantic supervision of category levels.
	\item \textbf{ZSIH (binary) \citep{shen2018zero}}  proposed a hashing model for the ZSSBIR task.
\end{itemize}
four SBIR methods
\begin{itemize}
	\item \textbf{Softmax (Baseline)} calculated the $4096\!-\!D$ (D stands for dimensional) VGG-16 feature vector for nearest neighbor search.
	\item \textbf{Siamese CNN \citep{qi2016sketch}} studied a novel Siamese network for coarse-grained SBIR.
	\item \textbf{SaN \citep{yu2015sketch}} constructed  a novel deep neural network, termed sketch-a-net, to capture the features of the sketches by employing the idea of AlexNet.
	\item \textbf{GN Triplet \citep{sangkloy2016sketchy}}  presented the first large-scale collection of sketch-image pairs: Sketchy database, and gave fine-grained relations between sketches and relevant images.
\end{itemize}
and five ZSL methods
\begin{itemize}
	\item \textbf{CMT \citep{socher2013zero}}  utilized outlier detection in the semantic space and two separate recognition models, which do not require manually defined semantic features for either words or images.
	
	\item \textbf{SSE \citep{zhang2015bit}}  proposed a new framework to generate bit-scalable hashing codes.
	
	\item \textbf{JLSE \citep{zhang2016zero}} formulated zero-shot learning as  a binary prediction problem, and constructed a joint discriminative learning framework via dictionary learning.
	
	\item \textbf{SAE \citep{kodirov2017semantic}}  presented a semantic auto-encoder based ZSL method.
	
	\item \textbf{FRWGAN \citep{felix2018multi} }  introduced a multi-modal cycle consistent constraint for the GAN training, which forces the generated visual features to reconstruct the original semantic features.
\end{itemize}

{\bf Evaluation Metrics}:
We utilize the same criterion described in previous work \citep{shen2018zero}, i.e., mean average precision (mAP) and precision considering top $100$ (Prec@100) to evaluate the ranked retrieval results.   Average precision is the  precision averaged across all values of recall between $0$ and $1$. In general, given a  sketch query, the average precision (AP) can be computed  as
$$AP(N)= \sum^N_{s=1} P(s)\delta_r(s),$$
for the top  $N$ ranked retrieval results,   where $P(s)$ is the precision at cutoff of $s$ images, and $\delta_r(s)$ stands for  the change in recall that happened between cutoff $s-1$ and cutoff $s$.  Thus, the mAP can be computed as
$$mAP =\frac1n\sum^n_{i=1} AP_i,$$
where $AP_i$ is the AP of class $i$, $n$ is the total number of classes.

\subsection{Implementing Details}
The proposed WAD-CMSN model is trained with RMSprop optimizer on Pytorch  with initial  training rate $1E-4$.  The input size of the image is $224\times224$. For feature extraction of both sketches and images,  the pre-trained VGG16 on ImageNet dataset (before the final pooling layer) is considered as a feature extractor, and the final extracted features have dimension $512$. Thus,  $512\!-\!D$ vector is accepted by the two generators $G_{\theta_{sk}}$ and $G_{\theta_{im}}$, which output $64\!-\!D$ features. The generators $F_{\hat\theta_{sk}}$ and $F_{\hat\theta_{im}}$ accept $64\!-\!D$ features and output $512\!-\!D$ vector. Moreover, Wikipedia is utilized to pre-train the  text-based model and  extract $300\!-\!D$ word vectors. Under the zero-shot scenario, the seen categories are utilized to construct the hierarchy embedding.  Hence, the hierarchical embedding for Sketchy (Extended) (TU-Berlin (Extended) resp.) dataset contains $354$ ($664$ resp.) nodes (\cite{miller1995wordnet}). At first, the word vectors constructed from the text-based approach and the hierarchical model are joined together, and play the role of semantic knowledge supervision via adversarial training. The decoder is introduced to obtain the $64-\!$D features.  During test, retrieval features are attained by feeding all the sketches and images into the proposed network.  Secondly,  the distance between the features of the sketch and those of the images is computed. Finally, top $100$ images are selected and the retrieval accuracy is calculated.

\begin{figure*}[!htb]
	\centering
	\subfigure[]{\label{mapv}\includegraphics[width=0.46\textwidth]{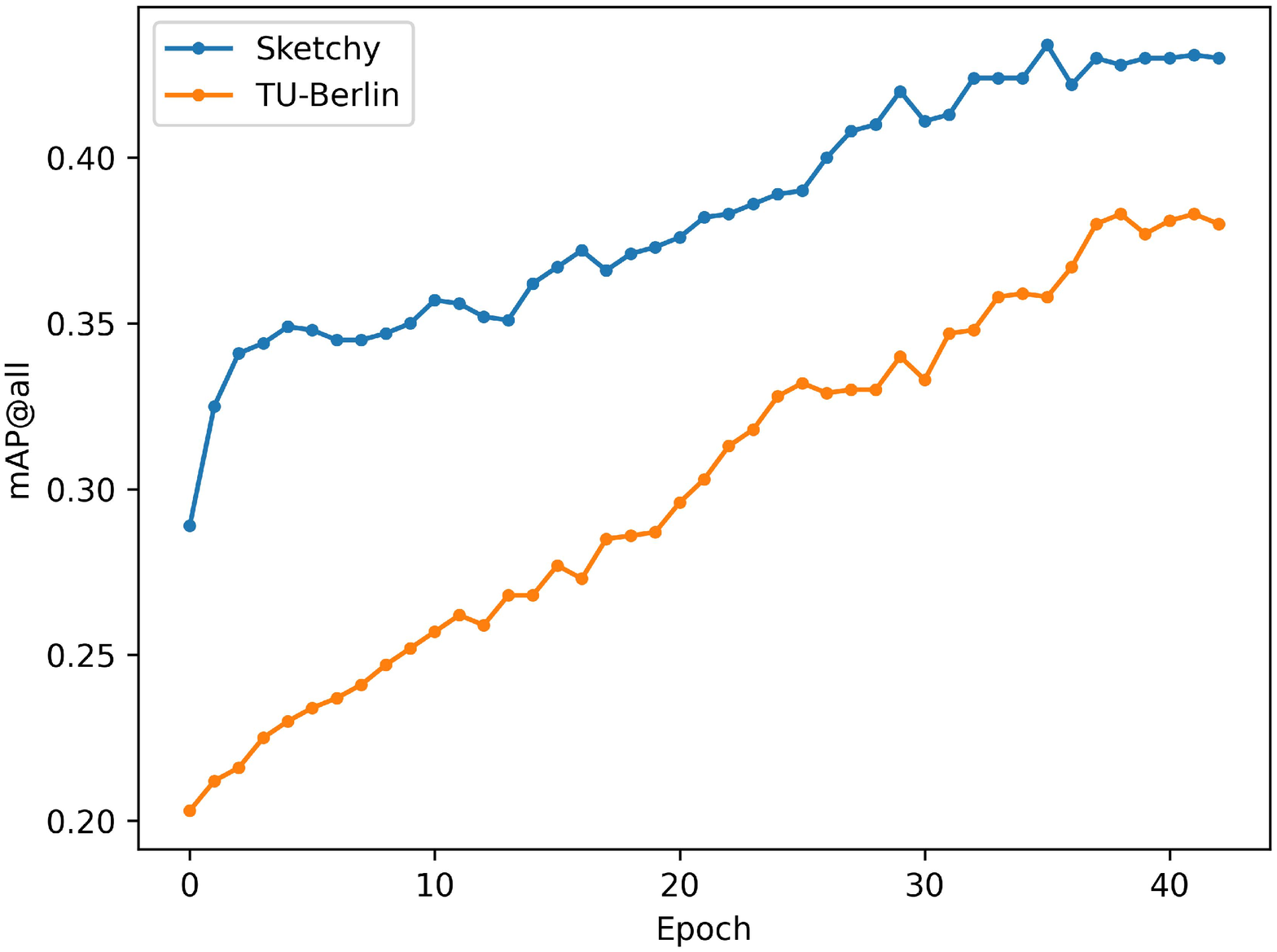}}
	\mbox{\hspace{0.1cm}}
	\subfigure[]{\label{lossv}\includegraphics[width=0.46\textwidth]{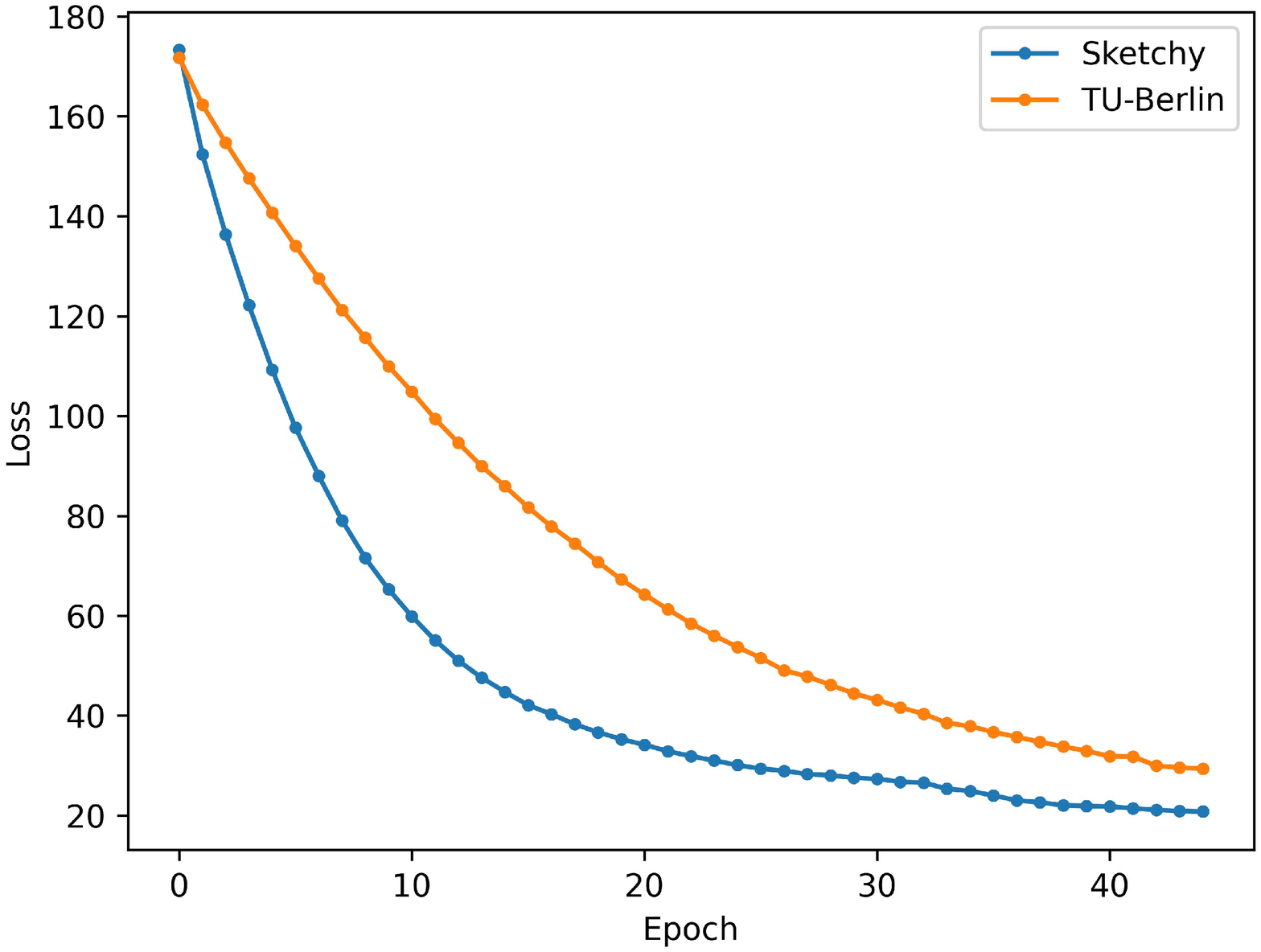}}
	\mbox{\hspace{0.1cm}}
\caption{The mAP value and loss value of each epoch for Sketchy (Extended) and TU-Berlin (Extended) dataset: (a). mAP  value; (b). loss value.}
	\label{map_epoch}
\end{figure*}

Furthermore, to conduct a fair comparison, we adopt the same validation set as described in  \cite{dutta2019semantically} to test the performance after training each epoch (takes about $70$ seconds). The mAP value and loss value of each epoch are  demonstrated in Fig. \ref{map_epoch}.

\subsection{Results}
The performance of all comparison methods is portrayed in Tables \ref{result_comp1} and \ref{result_comp2}. As can be observed from  Tables \ref{result_comp1} and \ref{result_comp2}, the proposed WAD-CMSN method outperforms all the SBIR, ZSL, and ZSSBIR methods.  This is because  SBIR  methods  aim at handling the problems of cross-modal retreival, and can not deal with knowledge transfer, while the ZSL  focuses on alleviating the semantic knowledge from seen categories to unavailable novel classes (unseen classes).  Therefore, the ZSSBIR methods,  which inherit the merits of SBIR and ZSL,  can achieve better performance. Moreover, under the zero-shot scenario, the proposed model outperforms the counterparts by at least $6\%$ ($4\%$ resp.) on Sketchy (Extended) (TU-Berlin (Extended) resp.) in terms of the mAP index. This indicates that the proposed model can efficiently bridge the significant domain gap, which gets benefits from the introduction of Wasserstein distance based adversarial loss, cycle consistency loss, identity matching loss and semantic knowledge supervision. We find that all the methods perform worse on TU-Berlin (Extended) database. This may be due to the overlapping and visual similarity in some categories of the TU-Berlin (Extended) database.

\begin{table*}[!htb]
	\caption {Comparison of retrieval performance  with competitors on Sketchy (Extended) dataset.}
	\label{result_comp1}
	\centering
	\scalebox{0.9}[0.9]{
	\begin{tabular}{ccccc}
		\hline
		\multicolumn{2}{c}{\multirow{2}{*}{Methods}} & \multicolumn{3}{c}{Sketchy (Extended)}\\
		&  &mAP &Precision@100 & Dimension	\\
		\hline
		\multirow{4}*{SBIR} &Softmax (Baseline) 	   &0.114	&0.172	&4096	\\	
		& Siamese CNN	       &0.132	&0.175	&64\\	
		&SaN	                  &0.115	&0.125	&512	\\	
		&GN Triplet 	           &0.204	&0.296	&1024	\\	
		\hline
		\multirow{5}*{ZSL} &CMT	   &0.087	&0.102	&300	\\	
		& SSE	       &0.116	&0.161	&100	 \\
		&JLSE                  &0.131	&0.185	&100	\\	
		&SAE	          &0.216	&0.293	&300	\\	
		&FRWGAN 	           &0.127	&0.169	&512\\		
		\hline
		\multirow{4}*{ZSSBIR} &CVAE 	   &0.196	&0.284	&4096	\\	
		&SEM-PCYC	       &0.349	&0.463	&64	 \\
		&ZSIH(binary)          &0.258	&0.342	&64		\\
		&\textbf{ours}	           &\textbf{0.415}	&\textbf{0.525}	&64		\\
		\hline
	\end{tabular}}
\end{table*}

\begin{figure*}[htb]
	\centering
	\includegraphics[scale=0.6]{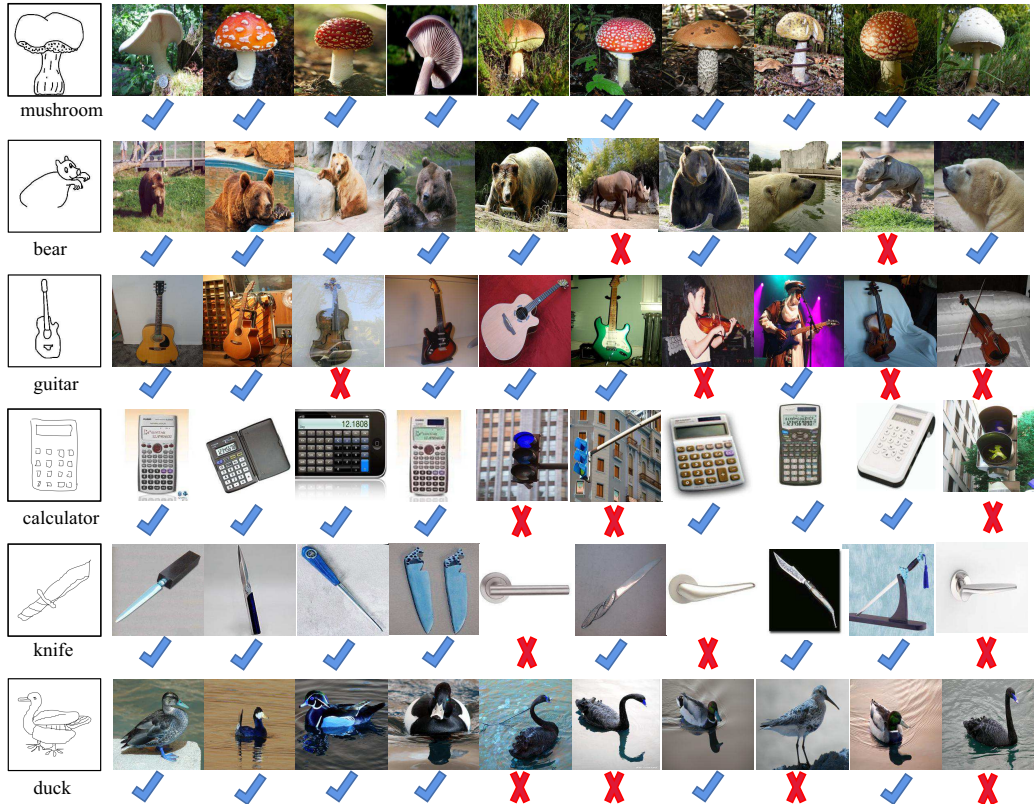}
	\caption{Top $10$ retrieved images attained by WAD-CMSN on Sketchy (Extendded) and TU-Berlin (Extended) datasets.}
	\label{Retrieve result}
\end{figure*}

We also analyze the retrieval performance of  our WAD-CMSN model qualitatively in Fig. \ref{Retrieve result}, where blue ticks mean correctly retrieved candidate, and red crosses denote wrongly retrieved candidate images.  As shown in Fig. \ref{Retrieve result}, sketch query of bear retrieves some rhinoceros, and querying knife retrieves some examples of door handle. This is probably because both of them are visually similar. Sketch query of guitar retrieves  some violins, probably because guitar and violin both belong to the instrument category, and they are semantic similar classes. In general, as can be observed, the wrongly retrieved candidate images and queried sketches mostly have closer visual or semantic relevance, which is more common in TU-Berlin (Extended) dataset. This may be due to the fact that  TU-Berlin (Extended) contains many categories of inter-class similar sketches. For instance, the duck and swan categories are visual similar, having the common watery background, and all belong to the bird category.

\begin{table*}[!htb]
	\caption {Comparison of retrieval performance  with competitors on TU-Berlin (Extended) dataset.}
	\label{result_comp2}
	\centering
	\scalebox{0.9}[0.9]{
	\begin{tabular}{cccccccc}
		\hline
		\multicolumn{2}{c}{\multirow{2}{*}{Methods}} 	& \multicolumn{3}{c}{TU-Berlin (Extended)}\\
		& &mAP &Precision@100 & Dimension		\\
		\hline
		\multirow{4}*{SBIR} &Softmax (Baseline) 	  	&0.089		&0.143     &4096\\
		& Siamese CNN 	          &0.109	    &0.141	    &64	\\
		&SaN	                  	&0.089	    &0.108	 	&512	\\
		&GN Triplet	  	&0.175	    &0.253	 	&1024	\\
		\hline
		\multirow{5}*{ZSL} &CMT  		&0.062		&0.078     &300\\
		& SSE     	    &0.089	    &0.121	    &220	\\
		&JLSE          &0.109	    &0.155	 	&220	\\
		&SAE	     &0.167	    &0.221	 	&300	\\
		&FRWGAN 	   	&0.110	    &0.157	 	&512	\\
		\hline
		\multirow{4}*{ZSSBIR} &CVAE 		&0.005		&0.001     &4096\\
		&SEM-PCYC	   &0.297	    &0.426	    &64	\\
		&ZSIH(binary)           	&0.223	    &0.294	 	&64	\\
		&\textbf{ours}	     	&\textbf{0.351}	    &\textbf{0.464}	 	&64	\\
		\hline
	\end{tabular}}
\end{table*}

\subsection{Effect of semantic knowledge supervision}
Semantic knowledge supervision, which provides semantic similarity between different categories is crucial in the zero-shot learning. We analyze and discuss the impacts of different semantic embeddings and their combinations, which are presented based on $64\!-\!\!D$ retrieval features. We demonstrated them in Table \ref{seman_embe}. As we can see, on Sketchy (Extended),  the combination of Word2vec and Jiang-Conrath's hierarchical similarity achieves the mAP at $0.415$, while on TU-Berlin (Extended) dataset, the combination of Word2vec and path similarity leads to the mAP value of $0.351$. We conclude from the experimental results that  text-based approach and hierarchal  model are crucial and complementary for ZSSBIR task.
\begin{table*}[!htb]\small
	\caption {The mAP results using different semantic embeddings (top) and their combinations (bottom).}
	\label{seman_embe}
	\centering
	\scalebox{0.8}[0.8]{
\begin{tabular}{cccccccccc}
	\hline
	\multicolumn{2}{c}{Text-based Model} & \multicolumn{2}{c}{Hierarchical Model}	&\multirow{2}{*}{Sketchy(Extended)} &\multirow{2}{*}{TU-Berlin(Extended)}\\
	 Word2vec  & Glove        &Path &Jiang-Conrath & & & &	\\
	\hline
	  &	    $\surd$   &	    &            	&0.294		&0.247	\\
     $\surd$      	 &  & 	   &    	        &0.346	    &0.254	    	\\
   	         &         &$\surd$&             &0.328		&0.256	    \\
	         &         &       & $\surd$     &0.331		&0.252	  \\
	\hline
	        &	  $\surd$    &$\surd$ &             &0.351		&0.331	\\
            &   	$\surd$     &        & $\surd$    &0.372	    &0.342	    	\\
  	 $\surd$ &    &$\surd$ &            &0.375     &\textbf{0.351}	    \\
     $\surd$ &   &        &$\surd$     &\textbf{0.415}		&0.345	  \\
	\hline
\end{tabular}}
\end{table*}

\subsection{Ablation Study}
\begin{figure*}[!htb]
	\centering
	\subfigure[]{\label{wwd1}\includegraphics[width=0.48\textwidth]{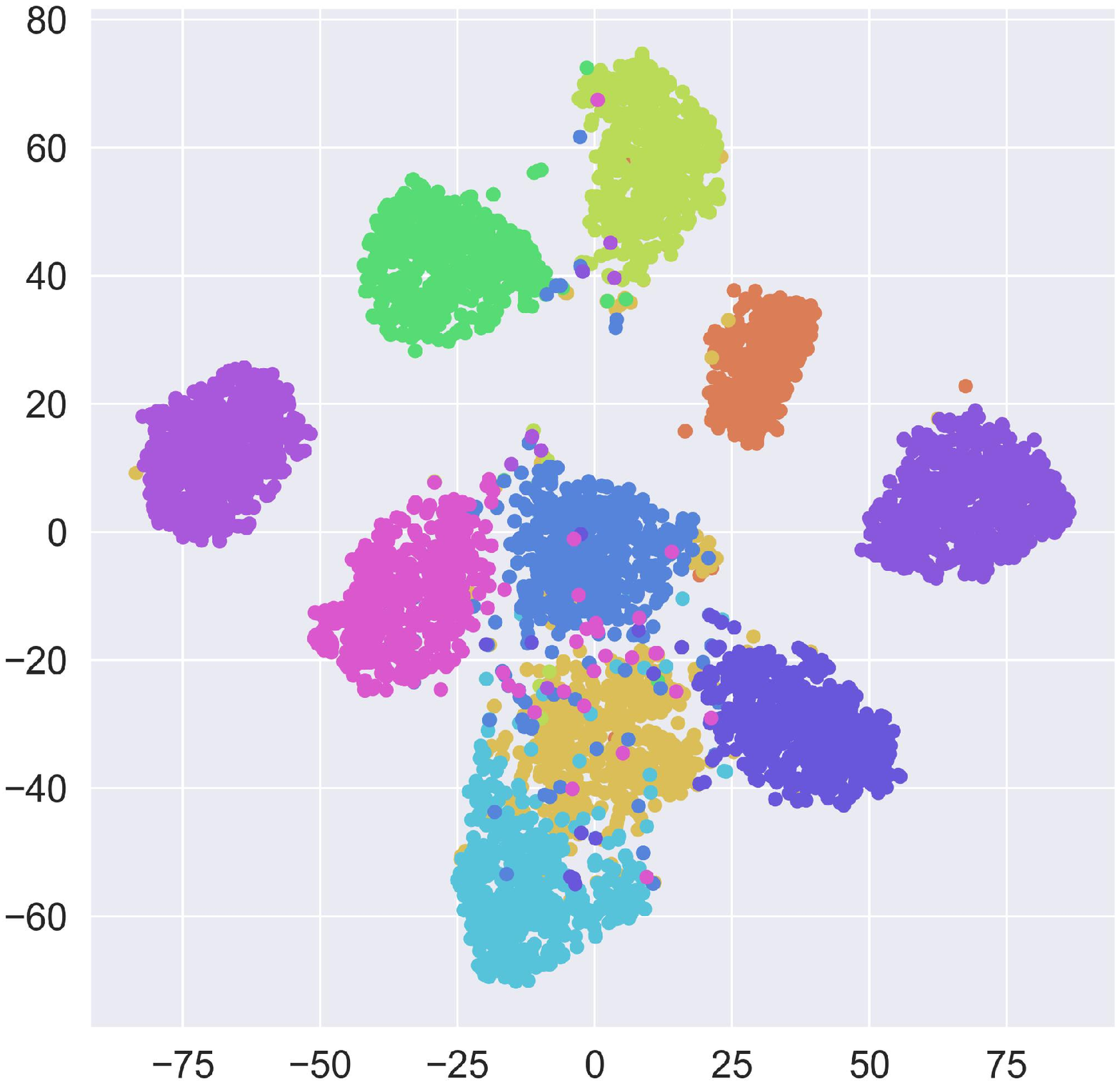}}
	\mbox{\hspace{0.1cm}}
	\subfigure[]{\label{wwd2}\includegraphics[width=0.48\textwidth]{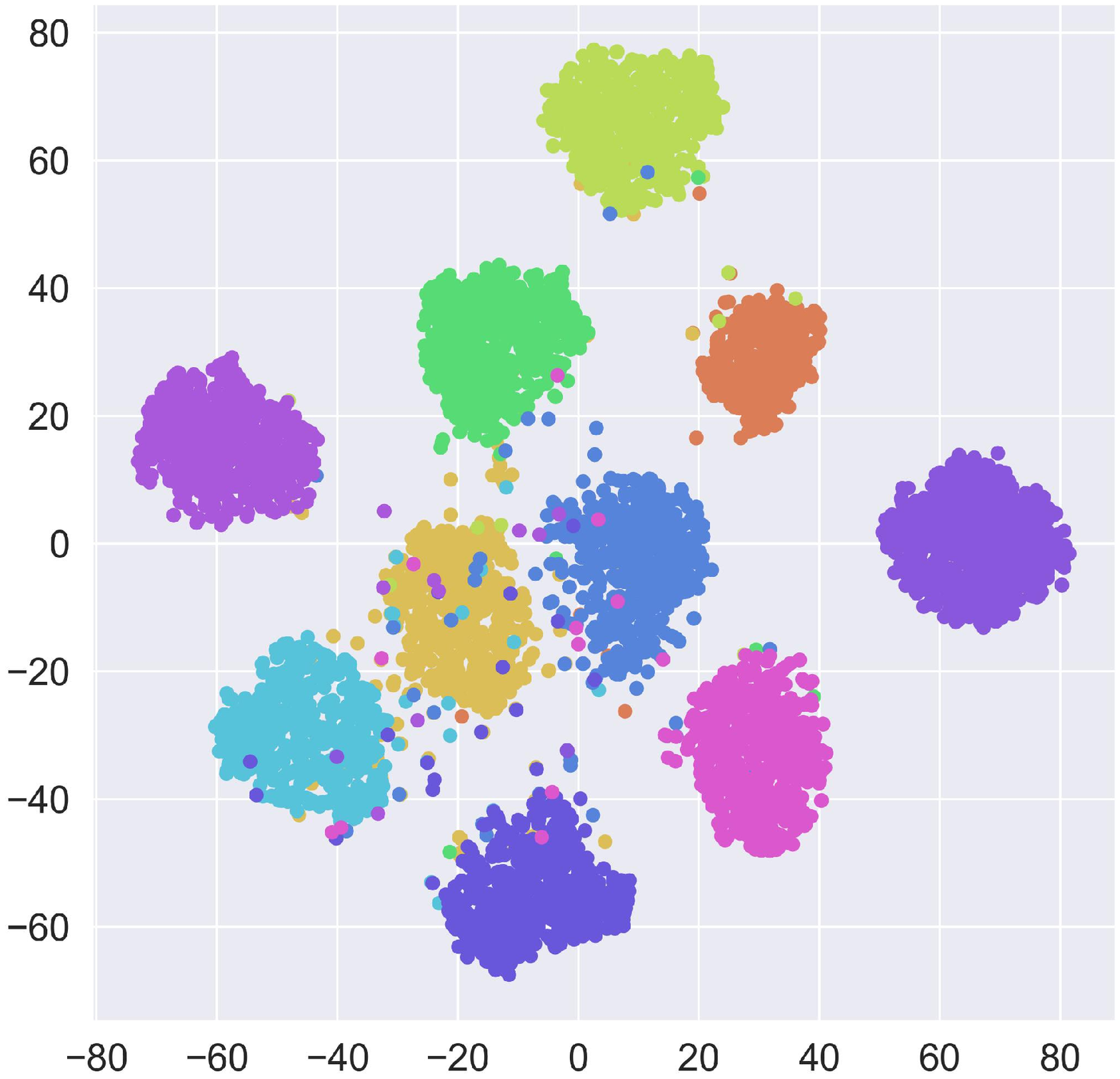}}
	\mbox{\hspace{0.1cm}}
	\caption{Visualization of the retrieval features on the TU-Berlin (Extended) dataset: (a). Without Wasserstein distance method; (b). Wasserstein distance based method}
	\label{wwwd_visu}
\end{figure*}

To justify the effectiveness of the proposed model,  several  ablation studies are investigated and displayed in Table \ref{ABLATION}. We first train a model that only uses adversarial loss. Secondly, to verify the effectiveness of the proposed method, we separately add identity matching loss and Wasserstein distance to test based on adversarial loss.  Thirdly, we set a baseline that considers Wasserstein distance based adversarial loss.  Finally, we add cycle consistency loss to the baseline,  and replace it with  classification loss and  identity matching loss alternatively.
\begin{table*}[!htbp]
	\caption {Ablation study in terms of the mAP values.}
	\label{ABLATION}
	\centering
	\scalebox{0.8}[0.8]{
\begin{tabular}{ccc}
	\hline
	Loss function &Sketchy (Extended)	&TU-Berlin (Extended)\\
	\hline
	Only adversarial loss           & 0.128       & 0.109\\
    Adversarial loss + identity matching loss & 0.241  & 0.199\\
	Adversarial loss+ Wasserstein distance(Baseline)     	&0.238		&0.207	\\
	Baseline+cycle consistency loss     &0.314	    &0.223	 \\
	Baseline+classification loss        &0.307	    &0.241	 \\
	Baseline +identity matching loss     &0.301		&0.232	  \\
	Our full model                    	    &\textbf{0.415}		&\textbf{0.351}\\
	\hline
\end{tabular}}
\end{table*}
The mAP values attained by different situations mentioned above are portrayed  in Table \ref{ABLATION}. As the results indicate, only considering  adversarial loss without Wasserstein distance, the performance of the proposed model drops significantly. When considering Wasserstein distance based adversarial loss, there is a great performance improvement about 11\% ( 9\%  resp.) on Sketchy (Extended) (TU-Berlin (Extended) resp.),  which means the Wasserstein distance based metric is reasonable and efficient for dealing with the feature gap issue.  Table \ref{ABLATION} also indicates that  the sketch and image from the same category are not well-matched when considering only adversarial loss. Therefore, we add different loss functions alternatively to the baseline. The performance of the model does not improve too much, however. Thus,  it can be concluded that all the four loss functions are complimentary to each other. Compared to the results without semantic embeddings, the performance of the full model improves significantly, which in turn demonstrates that semantic knowledge supervision is critical for the ZSSBIR task. Moreover, the distribution of the retrieved features  is visualized using t-SNE \citep{2008Visualizing} in Fig. \ref{wwwd_visu},  where $10$ classes (different colors in the figure) from sketch modality are selected using two distinct methods. As demonstrated in Fig. \ref{wwwd_visu}, Wasserstein distance based method yields discernible semantic features.

\section{Conclusion}\label{section5}
In this article, we presented a novel WAD-CMSN network for the ZSSBIR  task. Each branch of the proposed WAD-CMSN model maps a sketch and an image to a common low dimensional semantic space in an adversarial fashion. Firstly, cycle consistency loss imposed on both branches does not require the sketches and images to be aligned. Secondly, the classification loss on the generators' output ensures the features to be discriminative. Subsequently, identity matching loss can not only preserve the original features of the sketches and the images, but also alleviate the  over-fitting phenomenon. In addition, the use of semantic embeddings is important for  knowledge transfer. Experimental results  on two benchmarks: Sketchy (Extended) and TU-Berlin  (Extended) datasets indicate the efficacy of our WAD-CMSN model  compared to several existing ZSSBIR  methods. We believe the proposed model could be applied in other fields of artificial intelligence besides SBIR.


\bibliographystyle{abbrv}

\end{document}